\documentclass{article}

\usepackage{arxiv}

\usepackage[utf8]{inputenc} % allow utf-8 input
\usepackage[T1]{fontenc}    % use 8-bit T1 fonts
\usepackage{hyperref}       % hyperlinks
\usepackage{url}            % simple URL typesetting
\usepackage{booktabs}       % professional-quality tables
\usepackage{amsfonts}       % blackboard math symbols
\usepackage{nicefrac}       % compact symbols for 1/2, etc.
\usepackage{microtype}      % microtypography
\usepackage{graphicx}
\usepackage{float}
\usepackage{amsfonts}
\usepackage{mathtools}
\usepackage{amsmath}
\usepackage{amssymb}
\usepackage{todonotes}
\usepackage{bbold}

\title{
EPP: interpretable score of model predictive power\\
}

\author{
  Alicja Gosiewska \\
  Faculty of Mathematics and Information Science \\
  Warsaw University of Technology \\
  \texttt{alicjagosiewska@gmail.com} \\
  \url{https://orcid.org/0000-0001-6563-5742} \\
 \And
  Mateusz Bąkała \\
  Faculty of Mathematics and Information Science \\
  Warsaw University of Technology \\
 \AND
    Katarzyna Woźnica \\
  Faculty of Mathematics and Information Science \\
  Warsaw University of Technology \\
 \And
  Maciej Zwoliński \\
  Faculty of Mathematics and Information Science \\
  Warsaw University of Technology \\
  \AND
    Przemysław Biecek \\
 Faculty of Mathematics, Informatics and Mechanics\\ 
    University of Warsaw \\
  Faculty of Mathematics and Information Science\\
  Warsaw University of Technology\\
  \texttt{przemyslaw.biecek@gmail.com} \\
  \url{https://orcid.org/0000-0001-8423-1823} \\
}

\begin{document}
\maketitle

\begin{abstract}
The most important part of model selection and hyperparameter tuning is the evaluation of model performance. The most popular measures, such as AUC, F1, ACC for binary classification, or RMSE, MAD for regression, or cross-entropy for multilabel classification share two common weaknesses. First is, that they are not on an interval scale. It means that the difference in performance for the two models has no direct interpretation. It makes no sense to compare such differences between datasets. Second is, that for k-fold cross-validation, the model performance is in most cases calculated as an average performance from particular folds, which neglects the information how stable is the performance for different folds.

In this talk, we introduce a new EPP rating system for predictive models. We also demonstrate numerous advantages for this system, First, differences in EPP scores have probabilistic interpretation. Based on it we can assess the probability that one model will achieve better performance than another. 
Second, EPP scores can be directly compared between datasets. Third, they can be used for navigated hyperparameter tuning and model selection. Forth, we can create embeddings for datasets based on EPP scores.

\end{abstract}

\section{Introduction}

The measurement of performance is the foundation of the model selection and hyperparameter tuning. The choice of the algorithm strongly relies on the choice of an evaluation measure.
The task of evaluation score selection may be challenging, especially there are whole articles devoted only to the selection of various measures and their properties \citep{PowersD}.

For example, in binary classification problems, one of the most common measures is Area Under the ROC Curve (AUC) \citep{SOKOLOVA2009427}. However, if there is a high cost associated with False Negative examples, we would rather use Recall instead of AUC. If costs of False Positives is high we would use Precision. If we care about balance between precision and recall we should use F1 \citep{goutte2005probabilistic}.

In this paper, we will show weaknesses of the most common measures, such as AUC, F1, ACC for binary classification, or RMSE, MAE for regression, or cross-entropy for multilabel classification and propose our novel measure EPP (Elo-based Predictive Power).

The paper is organized as follows. In Section~\ref{sec:weaknesses} we present weaknesses of most common performance measures. In Section~\ref{sec:elo} we describe the idea of Elo score.
In Section~\ref{sec:epp} we introduce Elo-Based Predictive Power (EPP) measure and address all weaknesses pointed out in Section~\ref{sec:weaknesses}. In Section~\ref{sec:experiments} we show experiments and applications of EPP score. In Section~\ref{sec:extentions} we discuss possible extensions of EPP.

\section{What is wrong with most common measures?}
\label{sec:weaknesses}

In this section, we will point out four weaknesses of the most popular performance measures. We introduce examples for the AUC measure, however, reasoning would apply to other measures, such as F1, MSE, or cross-entropy. Each subsection correponds to a different issue.

\subsection{There is no interpretation of differences in performance}
\label{q:1}

\begin{table}[ht]
\centering
\begin{tabular}{ll}
Team                                   & AUC    \\ \hline
Erkut \& Mark,Google AutoML            & 0.618492 \\
Erkut \& Mark                          & 0.616913 \\
Google AutoML                          & 0.615982 \\
Erkut \& Mark,Google AutoML,Sweet Deal & 0.615858 \\
Sweet Deal                             & 0.615766 \\
Arno Candel @ H2O.ai                   & 0.615492 \\
ALDAPOP                                & 0.615040 \\
9hr Overfitness                        & 0.614371 \\
Shlandryn                              & 0.614132 \\
Erin (H2O AutoML 100 mins)             & 0.612657
\end{tabular}
\vspace{1em}
\caption{Top 10 results of KaggleDays SF competition in 2019.  \url{https://www.kaggle.com/antgoldbloom/analyzing-kaggledays-sf-competition-data/notebook}}
\label{tab:kaggle_automl}
\end{table}

In Table~\ref{tab:kaggle_automl} we present AUC of 10 machine learning models and AutoML solutions calculated on the same data set. 

The difference between AUC of the first and AUC of the second team equals $0.001579$. This difference has no direct interpretation, it does not provide any quantitative comparison of models' performances.  
AUC is useful for ordering, but its differences have no interpretation. 

\subsection{There is no procedure for assessing the significance of the difference in performances}
\label{q:2}
    
Results in the Table~\ref{tab:kaggle_automl} differ in the third decimal place. There is no reference point to indicate whether this difference represents a significant improvement in prediction or not. Significance in the statistical sense it means that these differences are not on the noise level.

\subsection{You cannot  compare performances between data sets}
\label{q:3}

\begin{table}[ht]
\begin{minipage}{0.5\linewidth}
\centering
\begin{tabular}{ll}
Team Name                              & AUC   \\ \hline
Asian Ensemble                         & 0.8043 \\
.baGGaj.                               & 0.8039 \\
Erkut \& Mark,Google AutoML,Sweet Deal & 0.8039 \\
ARG eMMSamble                          & 0.8037 \\
n\_m                                   & 0.8021
\end{tabular}
\caption{Springleaf Marketing Response Kaggle Competition, \url{https://www.kaggle.com/c/springleaf-marketing-response}}
\label{tab:kaggle_springleaf}
\end{minipage} 
\hfill
\begin{minipage}{0.45\linewidth}
\centering
\begin{tabular}{ll}
Team Name                      & AUC  \\ \hline
alijs                          & 0.9562 \\
7777777777777... & 0.9559 \\
ML Keksika                     & 0.9546 \\
krivoship                      & 0.9544 \\
2 old mipt dogs                & 0.9543
\end{tabular}
\caption{IEEE-CIS Fraud Detection Kaggle Competition, \url{https://www.kaggle.com/c/ieee-fraud-detection}}
\label{tab:kaggle_fraud}
\end{minipage}

\end{table}

In Tables~\ref{tab:kaggle_springleaf}~and~\ref{tab:kaggle_fraud} differences between best models for each data set are around $0.0003$. One would like to know, whether these differences are comparable between data sets. 
Does $0.0003$ on Springleaf Marketing data is the same increase of model quality on IEEE-CIS Fraud data? 

There at least three points of view. One is that the gap between the first and second places are almost the same for both data sets, because the differences in AUC are almost similar.
Second is that the gap in the IEEE-CIS Fraud Competition is larger as the AUC is close to 1. Relative improvement for Fraud detection 
($ \frac{0.9562 - 0.9559}{1- 0.9562} \approx 0.007$) 
is larger than relative improvement for Springleaf Marketing
($\frac{0.8043 - 0.8039}{1 - 0.8043} \approx 0.002$).
Third point of view is that the gap between first and second place for Springleaf is larger that the difference between second and third place. The opposite is true for IEEE-CIS Fraud detection.

\subsection{You cannot assess the stability of the performance in cross-validation folds}
\label{q:4}

\begin{table}[!ht]
\centering
\begin{tabular}{lll}
k        & AUC AutoML\_1 & AUC AutoML\_2 \\ \hline
1                 & 0.8                    & 0.9                       \\
2                 & 0.8                    & 0.78                    \\
3                 & 0.8                    & 0.78                    \\
4                 & 0.8                    & 0.78                   \\ \hline
\textbf{Mean AUC} & \textbf{0.8}           & \textbf{0.81}          
\end{tabular}
\caption{Artifficial results from 4-fold cross-validation.}
\label{tab:cv}
\end{table}

For k-fold cross-validation model performance is usually the averaged performance of models trained on different folds. In Table~\ref{tab:cv}, there are artificial values of AUC for four folds and mean AUC across all folds. 
Comparing just averages across folds creates false impression that the AutoML\_2 model is better than the AutoML\_1. Yet, we can see that AutoML\_1 wins in 3 out of 4 folds.

\section{What is Elo ranking system?}
\label{sec:elo}

 The Elo rating is a ranking system used for calculating the relative level of players' skill. It is used by, for example, chess and football federations. The score for players is updated after each match they have participated, new Elo rating is calculated on the basis of two components, result of match and rating of the opponent. 
 A player's level is not measured absolutely, although is inferred from wins, losses, and draws against other players. What is more, the difference between Elo scores of two players can be transferred into probabilities of winning when they play against each other.

 Elo scores can be interpret in terms of probability of winning. There are many variations of Elo, we will show a short overview of one of the most popular introduced by \citet{elo2008rating}.

 Let $S_1$ and $S_2$ be ratings of Player 1 and Player 2. The expected score of player 1 is
 $$E_1 = \frac{1}{1 + 10 ^{\frac{(S_1-S2)}{400}}}.$$ 
 Player 1 expected score $E_1$ is his probability of winning plus half of the probability of drawing with Player 2. After the match, the rating of Player 1 is updated using the formula given by:
 $$ S_1' = S_1 + K(A_1-E_1), $$
 where $A_1$ is actual score that means whether a player won or lost, it can take values 1 or 0. K is a given constant, which can take different values, it is usually defined by the organizer of the competition.
 
The most common scaling forces that the difference of 200 rating points mean that more skilled player has an expected score of approximately 0.75. An average player have a rating of 1500, and reaching rating over 2000 means that player is one of the best.
 
In addition to being interpretable in terms of probability, Elo has one more advantage. It is not necessary for each player to play with each other player. In real world, it would be impossible to play matches between all the chess players, therefore Elo is used to find an approximation of a true skill. Of course, the more matches played, the better approximation, however each player do not need to play with all other players.

The concept of Elo is not completely new in machine learning. The performance of neural networks that play Dota~2~are often expressed in terms of the TrueSkill which is a ranking system developed by  Microsoft \citep{herbrich2007trueskill} for e-sport. TrueSkill is an extension of Elo to games with more than two players. It is used to not only compare algorithms with each other, but also compare them with human players.
However, the Elo was not previously used to assess predictive models.

\section{Elo-based Predictive Power (EPP) score}
\label{sec:epp}
 
 Our novel idea is to transfer the way players are ranked in the Elo system to create rankings of models.
 
Let $\beta_{M_i}$ stands for EPP score for model $M_i$. The desired property is that 

\begin{equation}
\label{eq:log_reg_short}
\log (odds(i,j)) = \beta_{M_i} - \beta_{M_j}.
\end{equation}
where $odds(i,j)$ stands for odds that model $M_i$ beats model $M_j$.

% \subsection{Methodology}

% We introduce a new EPP (Elo-based Predictive Power) rating system for predictive models. 

% Let $ELO_{c}$ be Elo for model trained on $100-c$\% of data and validated on $c$\% of data. E.g. $ELO_{20}$ is our choice. It means that train\slash test split is  $80\%$\slash $20\%$.

The following procedure satisfies Property 1.

To calculate Elo we propose a logistic regression. 
Let $p_{i,j}$ be the probability of model $M_i$ wining with model $M_j$. Then we can specify formula

\begin{equation}
\label{eq:log_reg_short}
logit(p_{i,j}) = \beta_{M_i} - \beta_{M_j}.
\end{equation}
In case of larger number of models, it can be extended to 

\begin{equation}
\label{eq:log_reg}
logit(p_{i,j}) = \beta_{M_1} x_{M_1} + \beta_{M_2} x_{M_2} + ... + \beta_{M_k} x_{M_n} 
\end{equation}

where

\begin{equation}
  x_{M_a} =
  \begin{cases}
    1 & \text{if $a = i$} \\
   -1 & \text{if $a = j$} \\
    0 & \text{otherwise}
  \end{cases}
  .
\end{equation}

Unknown $\beta$ coefficients can be estimated with simple logistic regression.
Once $\beta$ coefficients are estimated, one can calculate $p_{i,j}$ from the following formula

\begin{equation}
\label{eq:prob}
p_{i,j} = invlogit(\beta_{M_i} - \beta_{M_j}) =  \frac{e^{\beta_{M_i} - \beta_{M_j} }}{1 + e^{\beta_{M_i} - \beta_{M_j}}}.
\end{equation}

\subsection{The advantages of EPP}

In this section, we will address four problems pointed out in Section~\ref{sec:weaknesses}, related to the weaknesses of the most common performance measures. We will show that EPP handles these identified issues.

\textbf{Ad \ref{q:1} There is an interpretation of differences in performance} \\
EPP score provides the direct interpretation in terms of probability. The EPP difference for models $M_i$ and $M_j$ is the logit of the probability that $M_i$ achieves better performance than $M_j$ (see Formula~\ref{eq:log_reg_short} and Formula~\ref{eq:prob}).

The next three points benefit from this probabilistic interpretation of the difference of scores.

\textbf{Ad \ref{q:2} There is a procedure for assessing the significance of the difference in performances} \\
EPP score allows to assess the significance via probability of better performance, which gives an intuition whether the difference in performance is a noise or not.

\textbf{Ad \ref{q:3} You can compare performances between data sets}\\
Difference between two EPP scores has he same meaning regardless of the data set. As stated in Equation~\ref{eq:log_reg_short}, it is logit of probability of better performance of one model over another.

\textbf{Ad \ref{q:4} You can assess the stability of the performance in cross-validation folds}\\
EPP score takes into consideration  how many times one model beat another. Thus, the better one would be the model that more often had higher performance.

\section{Experiments and applications of EPP score}
\label{sec:experiments}

\begin{figure}[htb]
\centering
    \includegraphics[width=0.85\textwidth]{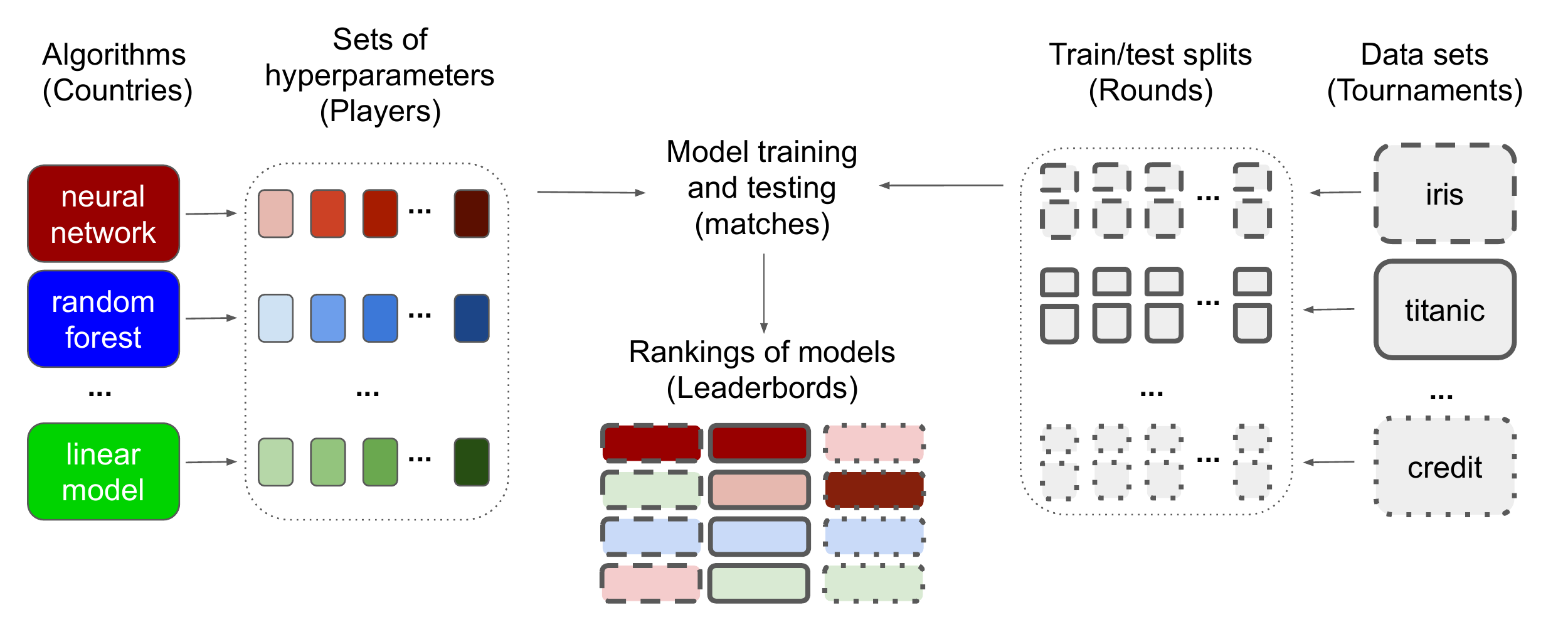}
    \caption{Our novel concept of Elo-based model ranking. Colors represent machine learning algorithms, gradients represent sets of hyperparameters, border styles represent data set.}
    \label{fig:elo}
\end{figure}

  In Figure~\ref{fig:elo}, we present a concept of Elo-based comparing of machine learning algorithms. 
We describe the ratings of models as an analogy to the tournaments with the Elo system. 
Countries (Algorithms) are staging their players (sets of hyperparameters) for duels. These duels are held within the tournaments (data sets) divided into rounds (train/test splits). The results of matches (models training and testing) are used to create leaderboards (EPP ranking of models).

The output rankings can be analyzed according to the type of algorithm, a specific set of hyperparameters, or particular data set.

We have calculated EPP score for several algorithms and data sets. In the following subsections we will present the results of the experiment.

\subsection{Experiment Setup}

We have used  4 machine learning algorithms (gradient boosting machines, generalized linear model with regularization, k nearest neighbours, and random forest). Each algorithm has been studied for 11 different hyperparameter settings on 11 selected classification data sets from the OpenML100 \citep{bischl2017openml} benchmark. 
For each data set, we specified 20 splits for train and test subsets. For each subset, we fitted models on train data and computed AUC on test data. 
For a model-data combination this gives us $11 \cdot 20 = 220$ values of AUC scores and the overall number of AUC values equals $220 * 4 * 11 = 9680$.

On the computed AUC scores, we applied methodology of calculating EPP presented in Section~\ref{sec:epp} and Figure~\ref{fig:elo}.~As a single round, we consider comparison of performances of two models with specified hyperparameters on the same data set, yet not necessary on the same train\slash test split.  
As a result, we have obtained EPP scores for each data-model-hyperparameters combination, which gave us $11 * 4 * 11 = 484$ values of EPP scores.

\subsection{Tuning hyperparameters of algorithms}

By analyzing EPP scores we can assess the tunability of model.  
\citet{JMLR:v20:18-444} had an attempt to measure the tunability of algorithms, EPP score can extend and add an interpretation for tunability measure. 

 \begin{figure}[htb]
\centering
    \includegraphics[width=0.85\textwidth]{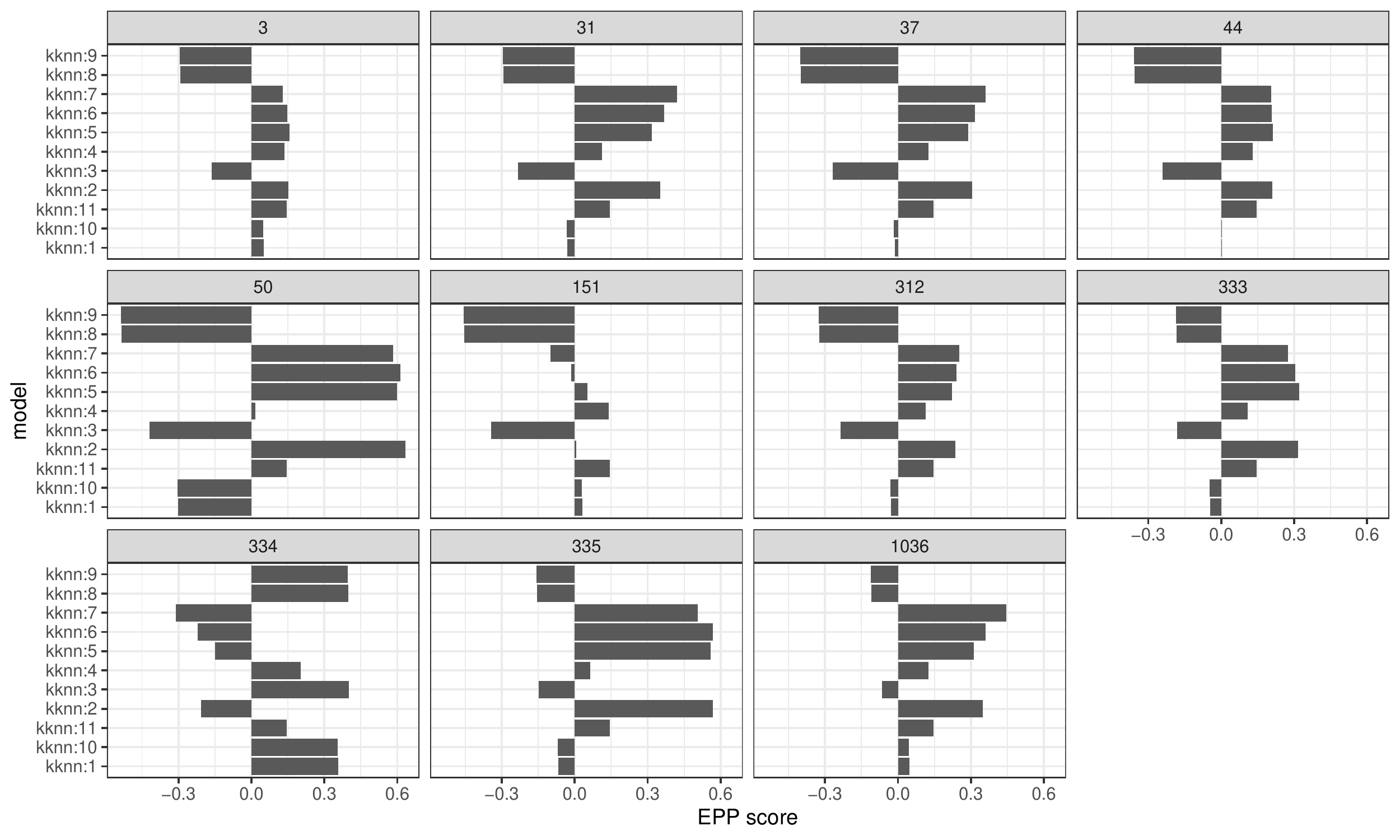}
    \caption{EPP scores of different hyperparameter settings for kknn across data sets. Each panel corresponds to the number of data set in the OpenML \citep{OpenML2013} data base. On the y-axis there are 11 different hyperparameter settings of k-nearest neighbours.}
    \label{fig:kknn}
\end{figure}

 \begin{figure}[!hbt]
\centering
    \includegraphics[width=0.8\textwidth]{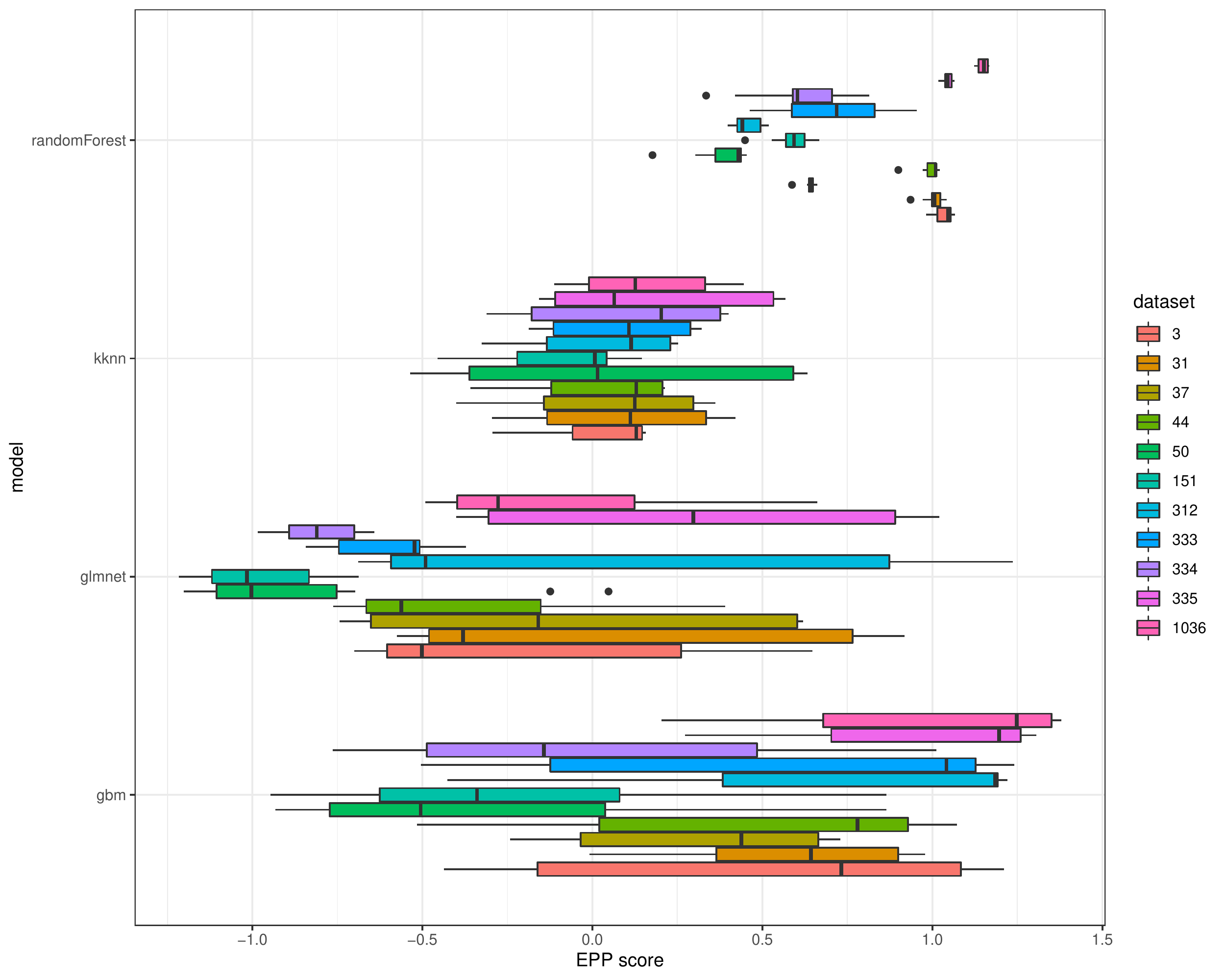}
    \caption{Boxplots of EPP scores for different models across data sets. }
    \label{fig:elo_boxplots}
\end{figure}

The EPP score has a huge potential for supporting hyperparameter tuning. In Figure~\ref{fig:kknn}, there are EPP scores for different hyperparameter settings of k-nearest neighbours across 11 data sets. The closer the end of the strap is to the right side, the better the model is. The further to the left side, the worse the model is. Models with EPP equals 0 have average~performances.

By looking on the results presented in Figure~\ref{fig:kknn}, we can make two insights. First, as the EPP scores differs, we can say that k-nearest neighbour model is susceptible to tuning. Second, data set number 334 is somehow different, as the direction of quality of models is reversed. During the process of modelling, it would be a hint that one should examine this data set.

In Figure~\ref{fig:elo_boxplots}, there are the distributions of EPP score across models and data sets. The longer boxplot is, the more tunable would be the model, for example, we can see that tree-based models (random forest and GBM) perform better on data set number 3 than the other two models. Also, all of EPP scores for random forest are positive, this means that generally, the performance of random forest is over the average.

The insights about the performance of models and particular hyperparameter settings could be further used for the navigated tuning that is an automated way to find the best model. 

\subsection{Building Embeddings of data sets}

  \begin{figure}[!htb]
\centering
    \includegraphics[width=0.85\textwidth]{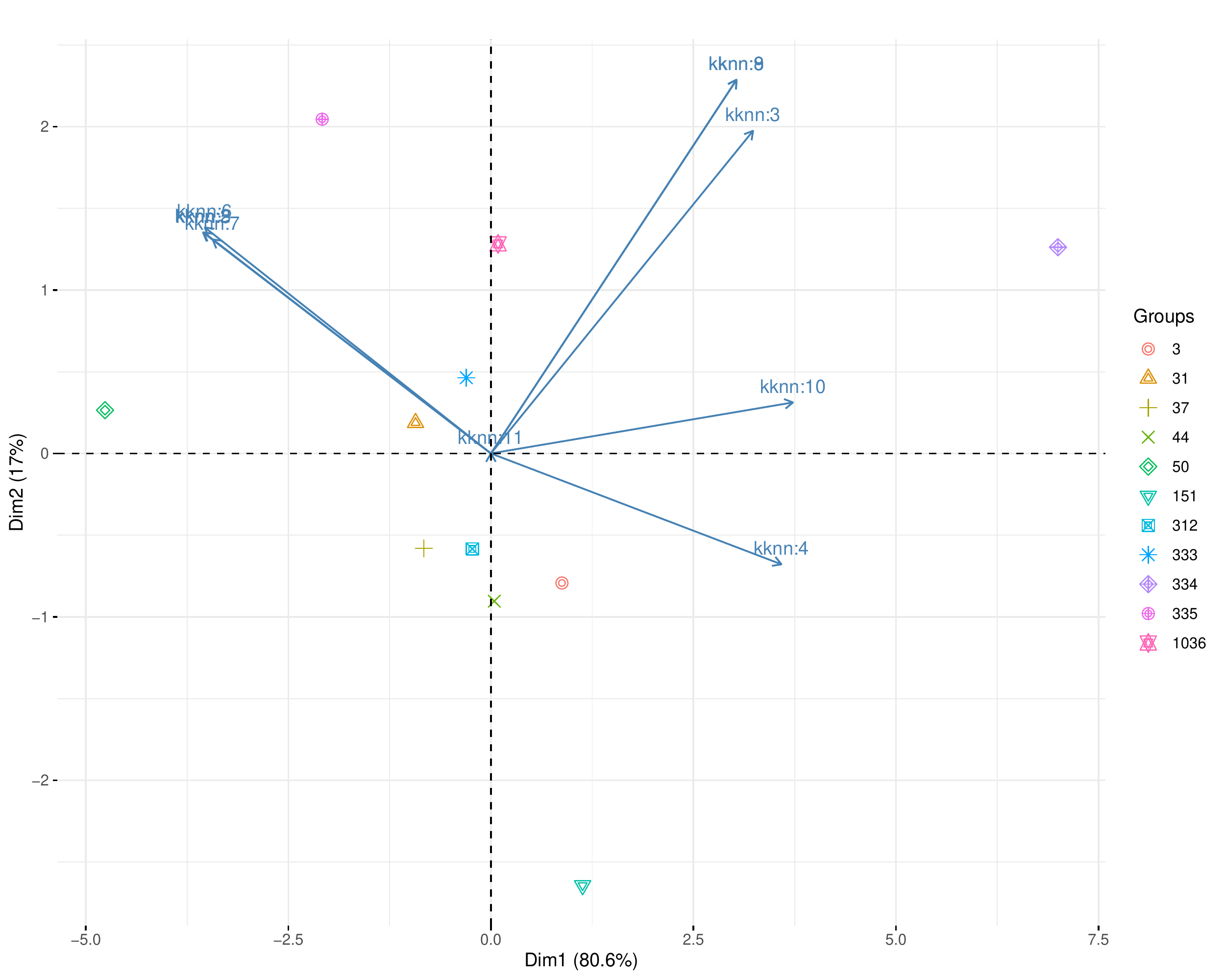}
    \caption{PCA biplot of EPP scores for different hyperparameter settings for k-nearest neighbours model. Groups presented as symbols correspond to the data sets.}
    \label{fig:pca}
\end{figure}

In the Figure~\ref{fig:pca},  there is a PCA biplot for different hyperparameter settings with marked datasets.  Such projection  can lead to additional insights.
We can observe a separation of hyperparameter settings for k-nearest neighbours.
The dimension linked with the x-axis provides a way to divide hyperparameter settings across data sets. Let us focus on the most marginal data sets, 50 and 334. When analyze results presented in bar plots in Figure~\ref{fig:kknn}, we can see that performances of individual hyperparameter settings are reversed for these two sets of data.
 
Due to the observation of a connection between the model and the data, one can use values of EPP to create embeddings of data sets. Such embeddings could be further used for model tuning.

\section{Possible extensions}
\label{sec:extentions}

The results of experiments are very promising. We see possible application and extentions in many areas of machine learning.

First of all, EPP score would be beneficial for Explainable Artificial Intelligence (XAI). Interpretability brings several
multiple benefits, such as, increasing trust in model predictions or identification of reasons behind poor predictions  \citep{JMLR:v19:18-416}. Interpretable differences of scores opens many new ways to develop explanations of machine learning~models.

The second major opportunity is to use EPP for navigated hyperparameter tuning. EPP score can be used to assess the probability that we can improve the performance if we continue searching of the hyperparameter space. What is more, the stop condition may also take into account the time of training further models. The automatization of the EPP-base tuning process could lead to developing navigated tuning method.

The idea od EPP score may by extended of a TrueSkill \citep{herbrich2007trueskill}, which was mentioned in Section~\ref{sec:elo}. The same way that TrueSkill allows to grade humans' skill in games for more than two players, it can be used for assessing the performance of model ensembles. It could make it possible to assess separately the performance of a single model, performance of the ensemble of models, and the potential of the model in the ensembles.

Possible modifications to the calculation of the EPP scores are also worth considering. For example, in experiments presented in Section~\ref{sec:experiments}, we have compared wins and loses of models across different train\slash test splits. An alternative way would be to compare the results only between identical splits.

\bibliographystyle{unsrt}
\bibliography{main}

\end{document}